\def\year{2020}\relax
\documentclass[letterpaper]{article} % DO NOT CHANGE THIS
\usepackage{aaai20}  % DO NOT CHANGE THIS
\usepackage{times}  % DO NOT CHANGE THIS
\usepackage{helvet} % DO NOT CHANGE THIS
\usepackage{courier}  % DO NOT CHANGE THIS
\usepackage[hyphens]{url}  % DO NOT CHANGE THIS
\usepackage{graphicx} % DO NOT CHANGE THIS
\usepackage{graphicx}  % DO NOT CHANGE THIS
\usepackage{amsmath}
\usepackage{systeme}

\title{Breaking Batch Normalization for better explainability of Deep Neural Networks through  Layer-wise Relevance Propagation }
\author{\Large \textbf{Mathilde Guillemot,\textsuperscript{\rm 1,2} Catherine Heusele, \textsuperscript{\rm 2} Rodolphe Korichi,\textsuperscript{\rm 2} Sylvianne Schnebert,\textsuperscript{\rm 2} Liming Chen \textsuperscript{\rm 1}}\\ % All authors must be in the same font size and format. Use \Large and \textbf to achieve this result when breaking a line
\textsuperscript{\rm 1}LIRIS - UMR 5205 CNRS\\ %If you have multiple authors and multiple affiliations
Ecole Centrale de Lyon\\
36, avenue Guy de Collongues, 69130 Ecully\\
\textsuperscript{\rm 2}LVMH Recherche - Parfums et cosmetiques\\
185 Avenue de Verdun, 45800 Saint-Jean-de-Braye\\
mathilde.guillemot@ec-lyon.fr\\
}
 
\begin{document}

\maketitle

\begin{abstract}

The lack of transparency of neural networks stays a major break for their use. The Layer-wise Relevance Propagation technique builds heat-maps representing the relevance of each input in the model's decision. The relevance spreads backward from the last to the first layer of the Deep Neural Network. Layer-wise Relevance Propagation does not manage normalization layers, in this work we suggest a method to include normalization layers. Specifically, we build an equivalent network fusing normalization layers and convolutional or fully connected layers. Heat-maps obtained with our method on MNIST and CIFAR-10 data-sets are more accurate for convolutional layers. Our study also prevents from using Layer-wise Relevance Propagation with networks including a combination of connected layers and normalization layer.
% Neural network show impressive results in many vision tasks. However, some, specifically in industrial world may be unwilling to exploit their potential because of their lack of transparency. 

% Understanding the way neural networks treat a problem, namely being able to link classification results and relevant entrance pixels, enable to learn new human knowledge or to improve the model. Therefore explainable artificial intelligence is an important research area. Layer-wise Relevance Propagation (LRP) is one out of several methods developed to handle neural network interpretability. LRP can be applied to fully connected or convolutional layers but does not manage normalization layers, specifically batch normalization layers. In this paper we propose a method based on the fusion of a normalization layer with closest fully connected or convolution layer. This method is tested on MNIST and CIFAR-10 data-sets with either fully connected neural network and convolutional neural networks. Models are run considering or bypassing normalization layer in the relevance calculation. Results point out that care should be taken when computing the relevance while using a fully connected network including batch normalization. For convolutional networks, adding batch normalization to the LRP method, results as significant improvement in the relevance heat maps.

\end{abstract}

Artificial intelligence methods are powerful and widely used in industry. %specifically in the research and development field. 
However, companies are more likely to use interpretable models even though they achieve lower performances. Consequently in an industrial context, companies prefer interpretable linear models to DNN considered as black boxes \cite{Lipton2016}. The innner working of these models is hard to understand, indeed DNN are complex models formed with multiple connections between neurons combined with non linear operations applied to all neurons. 
eXplainable Artificial Intelligence (XAI) develops new techniques to understand well performing known model rather than building interpretable models.

\citeauthor{Wojciech2017} (\citeyear{Wojciech2017}) mention that interpretability shows multiple interests. For delicate domains as autonomous car \citeauthor{Bojarski2017} or medical diagnosis \cite{Binder2018} a mistake from the model can have disastrous consequences. For such cases, before delivering any result, one must ensure that the classifier works as expected and takes decisions based on relevant information \cite{Stock2017}. Interpretability also enables to improve a classifier by addition of human experience i.e to analyze the learning algorithm's errors. The reverse operation, meaning learning from the model is a possibility thanks to interpretable artificial intelligence, opening really interesting perspective. For instance, the AlphaGo alorithm \cite{Silver2016} played during its game moves that a human cannot think of. Furthermore, the interpretability of neural network might lead to a better comprehension of the human brain as expected at the beginning of neural networks development \cite{Lettvin1959}. Finally the last benefit from XAI happens when neural networks achieve good results for a task a human cannot perform. Specifically in physics or chemical areas, interpretability would therefore enables to discover new principles as finding new genes linked to cancer or identify binding sites \cite{Schutt2017}.

\textbf{Contributions.} In this work, we propose an improvement of the Layer-Wise Propagation (LRP) method \cite{Bach2015}. LRP is a post-hoc interpretability method, implemented after the model training \cite{Bach2015} \cite{Marco2016}. It explains the model decisions one sample at a time. LRP propagates backward the relevance though all layers from output results to input features. The propagation follows different rules for convolutional layers, pooling layers, etc. However it is not clear how to manage normalization layers \cite{Montavon2017b}, and in some recent work, normalization layers are bypassed during the relevance back-propagation \cite{Montavon2017a}. We develop a method to easily include normalization layers to LRP method. We prove that properly fusing batch normalization(BN) \cite{Sergey2015} with another layer enables to integrate BN rather than bypassing it. We also show the extension of this method to other normalization layers. To assess the improvement brought by our method, it is tested on two data-sets MNIST \cite{Lecun1998} and  CIFAR-10 \cite{Krizhevsky2012}. Different networks architecture including fully connected neural networks\cite{Rumelhart1986} and convolutional neural networks \cite{LeCun1999} are tested. We demonstrate that relevance computation with BN obtains better results than ignoring BN for convolutional neural networks. Relatively to fully connected network, we reserve our conclusion since LRP seems not to be compatible with fully connected layers combined with batch normalization.

% For a company, the comprehension and explanation of a model is a serious challenge, the lack of interpretability of a model is a drag to the usage of such methods. 
 
%interpretation crucial on many practical application example + ref
\textbf{Related work.}
Several methods have been developed in order to deal with the explainable problem of DNN. First introduced methods build saliency maps \cite{Simonyan2013} or visualizations of patches that maximally activate neurons \cite{Girshick2013}. Other suggested gradient methods to explain reasons why images were misclassified \cite{Ramprasaath2016}. LIME \cite{Marco2016} or SHAP \cite{Lundberg2017} justify the predictions though an explainable classifier locally around the prediction.  Deep Taylor Decomposition \cite{Montavon2017a}, improves LRP, decomposing the activation of a neuron as the contributions from its outputs. DeepLIFT \cite{Shrikumar2018} decomposes the prediction by assigning the differences of contribution scores between the activation of each neuron to its reference activation. 

\section{Background and Notation}

We consider supervised learning tasks. Since LRP only processes samples one by one, notation of input features does not refer to the example index and is simply referred as $x^{(1)}$. The weights and biases between neuron i belonging to layer l $x_i^{(l)}$ and neuron j in to layer (l+1) $x_j^{(l+1)}$ are respectively written down $w_{ij}^{(l,l+1)}$ and $b_j^{(l,l+1)}$. Also $(.)^+$ means the ReLU function, i.e $max(.,0)$

\subsection{Fully connected layer}

Fully connected layers connect every neuron of one layer to every neuron of the next layer. Equation \ref{eq:6} gives the expression of the output neurons as a function of input neurons and of the fully connected layer parameters.

\begin{equation}\label{eq:6}
    \forall l, \forall i,  \forall j, x_j^{(l+1)} = \sum_i{w_{ij}^{(l,l+1)}x_i^{(l)}} + b_j^{l}
\end{equation}

\subsection{Convolutional Layer}
Unlike fully connected layers, convolutional layers contain a set of filters.
Each filter is convolved with the input layer (l) to compute an activation map. The filter is slid across the width and height of the input and the dot products between the input and filter are computed at every spatial position \ref{eq:9}.

\begin{equation}\label{eq:9}
    \begin{aligned}
        x^{(l+1)} = w^{(l,l+1)}*x^{(l)} + b^{(l,l+1)}\\
    \end{aligned}
\end{equation}

\subsection{Batch Normalization}
Batch normalization \cite{Sergey2015}, is a trick commonly used to improve the training of deep neural networks, accelerating learning phase and showing better accuracy. Its success leads various deep learning structure to incorporate batch normalization \cite{Kaiming2015} \cite{Gao2016}. During the learning phase, batch normalization avoids problems related to back-propagation. It prevents the gradient from explosion and the vanishing gradient problem by keeping data in bounded intervals. 
During the test phase, batch normalization is performed using constant variance and constant mean. 
Equation \ref{eq:5} expresses the output ${x_i^{(l)}}_{norm}$ of a BN layer $(l)$ as a function of the input $x_i^{(l)}$ during the test phase.

\begin{equation}\label{eq:5}
{x_i^{(l)}}_{norm} = \gamma^{(l)} \frac{x_i^{(l)}-\mu_{run}^{(l)}}{\sigma_{run}^{(l)}} + \beta^{(l)} 
\end{equation}
with $\gamma^{(l)}$ and $\beta^{(l)}$ respectively weights and biases of the layer.

\subsection{Layer-wise Relevance Propagation}

\subsubsection{Method}
LRP performs once the network is learned. It suggests to find the relevance $R_i$ of each input feature $x_i^{(1)}$, propagating backward the relevance information from the output until the input. 
The relevance obeys to conservation rule from one layer to another \ref{eq:1}.
\begin{equation}\label{eq:1}
    \forall l, \sum_i R_i^{(l)} = \sum_j{R_j^{(l+1)}}
\end{equation}
Equation \ref{eq:2} translates $R_i^{(l)}$ , the relevance of neuron i in layer l, as the sum of all the contribution of neuron communicating with it.
 \begin{equation}\label{eq:2}
    \forall l, \forall i, R_i^{(l)} = \sum_j{R_{i\leftarrow j}^{(l+1)}}
\end{equation}

\citeauthor{Montavon2017b} (\citeyear{Montavon2017b}), establishe rules satisfying equations \ref{eq:1} and \ref{eq:2} to propagate the relevance from a layer to the previous one. In this paper we will use two of them depending on the input domain.
\begin{itemize}
    \item Rule 1 : If the neuron value $x_i^{(l)}$ is positive. The relevance of the neuron i of the layer l is computed as in equation \ref{eq:3}
     \begin{equation}\label{eq:3}
        {R_i}^{(l)} = \sum_j{\frac{{x_i}^{(l)} {{w_{ij}}^{(l,l+1)}}^+}{\sum_k {x_k}^{(l)} {{w_{kj}}^{(l,l+1)}}^+}{R_j}^{(l+1)}}
     \end{equation}
     with $ {w_{ij}^{(l,l+1)}}^+ = max(0,  w_{ij}^{(l,l+1)})$
     \item Rule 2 : If the neuron values $x_i^{(l)}$ range between $l_i$ and $h_i$
     \begin{equation}\label{eq:4}
     \scriptsize{
     \begin{aligned}
        &{R_i}^{(l)} = \sum_j{k_ijl} {{R_j}^{(l+1)}}\\
        &k_{ijl} = {\frac{x_i^{(l)} {w_{ij}}^{(l,l+1)} - {l_i}^{(l)}  {{w_{ij}}^{(l,l+1)}}^+ - {h_i}^{(l)} {{w_{ij}}^{(l,l+1)}}^-  }{\sum_k{{x_k}^{(l)} {w_{kj}}^{(l,l+1)}- {l_k}^{(l)}  {{w_{kj}}^{(l,l+1)}}^+ - {h_k}^{(l)} {{w_{kj}}^{(l,l+1)}}^-}}}
        \end{aligned}}%
     \end{equation}
     with $ {{w_{ij}}^{(l,l+1)}}^+ = max(0,  {w_{ij}}^{(l,l+1)})$ and $ {{w_{ij}}^{(l,l+1)}}^- = min(0,  {w_{ij}}^{(l,l+1)})$
     
\end{itemize}

Relevance R computed with LRP is pictured as a heat-map. The heat-map is a visualization technique highlighting pixels which support the classification decisions \cite{Simonyan2013} \cite{Bach2015}.

\subsubsection{Practical Considerations}
The input of LRP corresponds to the raw output of the network i.e before softmax activation function.
Next, the chosen non linear activation function is a ReLU. As a consequence, for all $l \ne 0$ and for all i, $x_i^{(l)} > 0$  and the relevance is propagated according to equation \ref{eq:3}. 
As we work with images, pixel inputs $x_i^{(0)}$ are bounded between 0 and 255 (or -1 and 1 if a scale operation is applied) and  rule 2 \ref{eq:4} is applied.

Pooling layers are easily handled, being considered as reLU detection layers.

\citeauthor{Montavon2017b} (\citeyear{Montavon2017b}) recommend all biases to be either zero or negative. When the condition is not filled, biases are considered as neurons and their contribution is added on the denominator of equation \ref{eq:3} or \ref{eq:4} \cite{Montavon2017a}.

All results shown  are part of the test set.

\section{Batch normalization in the LRP relevance computation}
Various pre-trained networks \cite{Szegedy2015} \cite{Howard2017} \cite{Gao2016} include normalization layers and show good results on various tasks. Such networks support LRP with approximation, but we expect better results with a suitable way to handle BN layers with LRP. We propose a new method to obtain the relevance heat map of a DNN classifier with BN layers.

\begin{figure}[t]
\includegraphics[width=8cm]{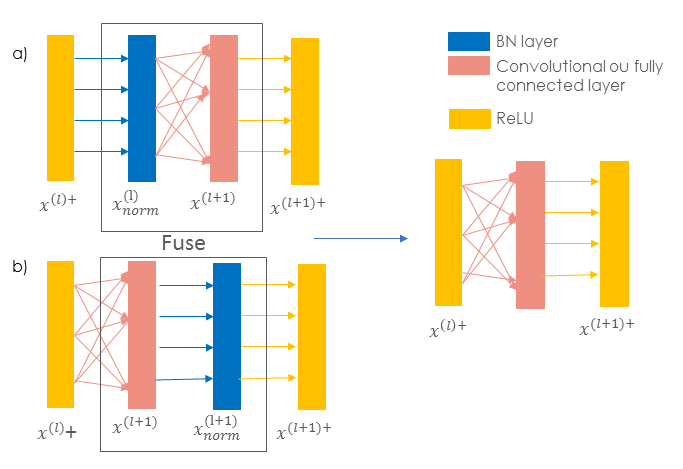}
\caption{Fusion of the normalization layer with another layer in the two different configurations a) BN after activation b) BN before activation}
\label{fig:0}
\centering
\end{figure}

The normalization layer is applied, before or after the activation. The general idea of our method is to fuse the batch normalization layer with the closest convolutional or fully connected layer (see figure \ref{fig:0}) into a single convolutional or fully connected layer simply by modifying their parameters. ${{w_{ij}}^{(l,l+1)}}'$ and ${{b_{ij}}^{(l,l+1)}}'$ are the parameters of this new layer.

\subsection{Fully connected neural network}
% Either way, with the notation defined in section \nameref{notations}, a batch normalization applied after the layer $l$ to the neurons $x$ is defined as the equation \ref{eq:5}. 

% \begin{equation}\label{eq:5}
%     \forall l,  \forall i,  x_i^{(l)}_{norm} = \gamma_i^{(l)} \frac{x_i^{(l)}-\mu_i^{(l)}_{run}}{\sigma_i^{(l)}_{run}} + \beta_i^{(l)}
% \end{equation}

% Considering a fully connected connection between two neurons layers, the neurons belonging to layer (l+1) can be express as a function of the parameters of the fully connected layer, weight matrix $w^{(l,l+1)}$ and bias vector $b^{(l,l+1)}$ see equation \ref{eq:6}.

% \begin{equation}\label{eq:6}
%     \forall l, \forall i,  \forall j, x_j^{(l+1)} = \sum_i{w_{ij}^{(l,l+1)}x_i^{(l)}} + b_j^{l}
% \end{equation}

Combining the two equations \ref{eq:6} and \ref{eq:5}, we show that the combination of a BN layer and a FC layer is equivalent to a single fully connected layer by adapting its weight and bias parameters.

\subsubsection{BN after activation} (see \ref{fig:0}, graph a))

If a BN fuses with a FC layer, the combination is written as equation \ref{eq:7}.
\begin{small}
\begin{equation}\label{eq:7}
    \begin{aligned}
        \forall l, \forall j, {x_j}^{(l+1)} = {} & \sum_i{{w_{ij}}^{(l,l+1)} \left(  \gamma_i^{(l)} \frac{{x_i}^{(l)}-{{\mu_i}^{(l)}}_{run}}{{{\sigma_i}^{(l)}}_{run}} + {\beta}^{(l)} \right)}\\ & + {b_j}^{(l,l+1)}
        \end{aligned}
\end{equation}
\begin{equation}\label{eq:7bis}
    \begin{aligned}
    \forall l, \forall j, {x_j}^{(l+1)} = \sum_i{\left[ \frac{{\gamma_i}^{(l)}{w_{ij}}^{(l,l+1)}}{{{\sigma_i}^{(l)}}_{run}}\right] {x_i}^{(l)}} \\ + \left[ {b_j}^{(l,l+1)} + \sum_i{{w_{ij}}^{(l,l+1)} \left( \beta^{(l)}- \frac{{\gamma_i}^{(l)}{{\mu_i}^{(l)}}_{run}}{{{\sigma_i}^{(l)}}_{run}}\right)} \right]
    \end{aligned}
\end{equation}
\end{small}
By identification, we find weight and bias terms of the fuse layer respectively in the first and second bracket of equation \ref{eq:7bis}. 

% It is possible to replace the combination of a BN and a FC by a single FC layer with the following parameters : 
% \begin{itemize}
%     \item $w_{ij}^{(l,l+1)'}$ = $\frac{\gamma_i^{(l)}w_{ij}^{(l,l+1)}}{\sigma_i^{(l)}_{run}}$ 
%     \item $b_{j}^{(l,l+1)'}$ = $ b_j^{l,l+1} +\sum_i{w_{ij}^{(l,l+1)}(\beta_i^{(l)} - \frac{\gamma_i^{(l)}\mu_i^{(l)}_{run}}{\sigma_i^{(l)}_{run}})}$
% \end{itemize}

\subsubsection{BN before activation} (see \ref{fig:0}, graph b))

Similarly, we find out that a FC layer and a BN layer can be fused into a single FC layer with the following parameters.
\begin{align*}
     & {{w_{ij}}^{(l,l+1)}}' = \frac{{\gamma_j}^{(l+1)}{w_{ij}}^{(l,l+1)}}{{{\sigma_j}^{(l+1)}}_{run}} \\
     & {b_{j}^{(l,l+1)}}' = {\beta_j}^{(l+1)} + {{\gamma_j}^{(l+1)}\frac{{b_j}^{(l,l+1)} - {{\mu_j}^{(l+1)}}_{run}}{{{\sigma_j}^{(l+1)}}_{run}}} 
\end{align*}

\subsection{Convolutional neural network}
\begin{figure}
    \centering
    \includegraphics[width=8cm]{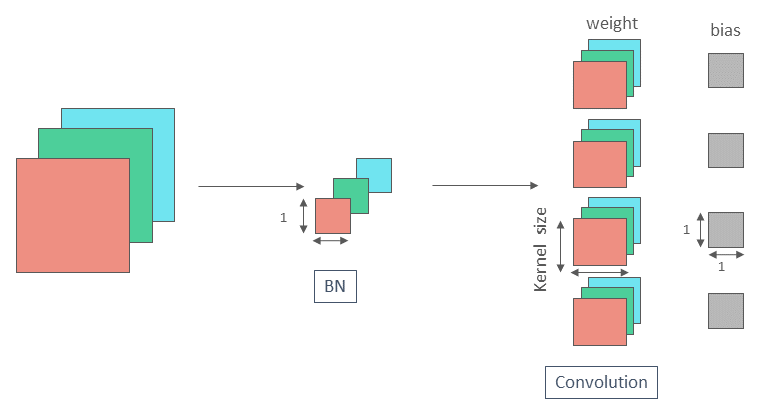}
    \caption{Scheme of a batch normalization applied before a convolutional layer}
    \label{fig:1}
\end{figure}

For the batch normalization combined with a fully connected layer, all input neurons receive a different normalization i.e $\gamma^{l}$, $\beta^{(l)}$, $\sigma^{(l+1)}_{run}$ and $\mu^{(l)}_{run}$ are vector which size is equal to the neurons vector's one. With a convolutional layer,  the same normalization is applied to all neurons (see Figure. \ref{fig:1}). BN parameters are simplified :
\begin{align} \label{simplifications} 
    \forall i,  &{\gamma_i}^{(l)} =  \gamma^{(l)} \nonumber \\ 
    &{\beta_i}^{(l)} =\beta^{(l)}\\
    &{{\mu_i}^{(l)}}_{run}= {mu^{(l)}}_{run} \nonumber \\ 
    &{{\sigma_i}^{(l)}}_{run} = {\sigma^{(l)}}_{run} \nonumber 
\end{align}
As a consequence, we rewrite equation \ref{eq:5} removing unnecessary terms.

% \begin{equation}\label{eq:8}
%     \forall l,  \forall i,  x_i^{(l)}_{norm} = \gamma^{(l)} \frac{x_i^{(l)}-\mu^{(l)}_{run}}{\sigma^{(l)}_{run}} + \beta^{(l)}
% \end{equation}

% Moreover for a convolutional network, the link between neurons of layer (l+1) and neuron belonging to layer (l) is explained in equation \ref{eq:9}. 

% \begin{equation}\label{eq:9}
%     \begin{aligned}
%         x^{(l+1)} = w^{(l,l+1)}*x^{(l)} + b^{(l,l+1)}\\
%     \end{aligned}
% \end{equation}

\subsubsection{BN after activation}

Using equation \ref{eq:9} and \ref{eq:5} and proceeding as in the previous sections, a convolutional layer combined with a BN layer can be reduced to a single convolution layer which weight and bias are expressed as : 
\begin{align*}
    & {w^{(l,l+1)}}' = \frac{ w^{(l,l+1)}\gamma^{(l)}}{{\sigma^{(l)}}_{run}} \\
    & {b^{(l,l+1)}}' = b^{(l,l+1)} + w^{(l,l+1)}(\beta^{(l)} - \frac{\gamma^{(l)}{\mu^{(l)}}_{run}}{\sigma^{(l)}}
\end{align*}

\subsubsection{BN before activation}
Similarly if we apply BN before non linearity, the weights and bias of the resulting fused convolutional layer are expressed as :

\begin{align*}
    & w^{(l,l+1)'} = \frac{ w^{(l,l+1)}\gamma^{(l)}}{\sigma^{(l)}_{run}}\\
    & b^{(l,l+1)'} = \beta^{(l+1)} + \gamma^{(l+1)} \frac{b^{(l,l+1)} - \mu^{(l+1)}_{run}}{\sigma^{(l+1)}}
\end{align*}

\section{More complex normalizations for convolution layers - from convolutional to fully connected layer}
Batch normalization at test time consists in applying the same mean and variance for all coefficients of an input (see figure \ref{fig:1}). For other normalizations, parameters (mean, variance etc) may not be constant. 
The simplifications  expressed in equation \ref{simplifications} are no longer practical. An other way to proceed must be found. 

% \subsection{Fully connected Layer}
% There is no change for the fully connected layer used in combinaison with batch normalization. In this case, the batch normalization parameters have the same size as the input vector. Same formulas can be used. 

\begin{figure}
    \centering
    \includegraphics[width=8cm]{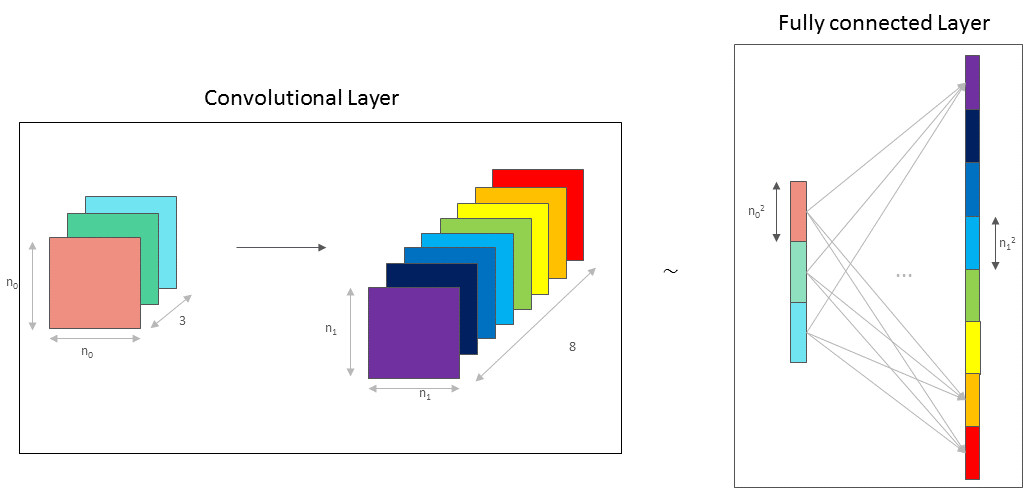}
    \caption{Example of a convolution layer and its equivalent as a fully connected layer after  a flatten operation}
    \label{fig:2}
\end{figure}

We thought to reduce the convolutional layer to a fully connected layer with weight $W_{fc}$ and bias $B_{fc}$ and then apply the known results on FC.
In practice two dimensional inputs and outputs of the layer are flatten (see figure \ref{fig:2}). And a weight matrix of the new created fully connected layer is filled with the coefficients of the different kernels. 

For each connection between an input flatten channel and an output flatten channel, a weight matrix is created. Given $c_0$, $c_1$ respectively the number of input and output channels, for all i $\in [1,c_0]$ and for all j $\in [1,c_1]$, a matrix weight $w_{ij}$ and a bias vector $b_j$ are filled. The final weight matrix and bias vector are the concatenation of those sub-matrix :
\newline

\begin{small}
$W_{fc}=\begin{pmatrix}
  w_{1,1} & w_{1,2} & \cdots & w_{1,c_0} \\
  w_{2,1} & w_{2,2} & \cdots & w_{2,c_0} \\
  \vdots  & \vdots  & \ddots & \vdots  \\
  w_{c_1,1} & w_{c_1,2} & \cdots & w_{c_1,c_0} 
 \end{pmatrix}$
and 
 $B_{fc} =  \begin{pmatrix}
  b_{1} \\
  b_{2} \\
  \vdots \\
  b_{c_1,} 
 \end{pmatrix}$
\end{small}

The matrix $w_{ij}$ and vector $b_j$ coefficients are expressed thanks to kernel coefficient of the concerned couple of input channel i and output channel j. For instance considering a convolutional layer with an input channel with 5*5 images, a 2*2 kernel, no padding and no stride. The coefficients of the kernel are written $\alpha_i$  and the pixels are noted $p_i$. The convolution can be drawn as :

\begin{center}
$W_{fc}=\begin{pmatrix}
  p_1 & p_2 & \cdots & p_5 \\
  p_6 & p_7 & \cdots & p_10 \\
  \vdots  & \vdots  & \ddots & \vdots  \\
  p_21 & p_22 & \cdots & p_25 
 \end{pmatrix} * \begin{pmatrix} \alpha_1 &\alpha_2\\
 \alpha_3 &\alpha_4\\
 \end{pmatrix}$
 \end{center}

In this case, the sub-matrix weight matrix $w_{ij} $ is 
\newline
$w_{ij} =\begin{pmatrix}
  \alpha_1 & \alpha_2 & 0 & 0 & 0 & \alpha_3 & \alpha_4 & 0 & \cdots & 0 \\
   0 & \alpha_1 & \alpha_2 & 0 & 0 & 0 & \alpha_3 & \alpha_4 & \cdots & 0 \\
   0 & 0 & \alpha_1 & \alpha_2 & 0 & 0 & 0 & \alpha_3 & \cdots & 0 \\
   0 & 0 & 0 & \alpha_1 & \alpha_2 & 0 & 0 & 0 & \cdots & 0 \\
   0 & 0 & 0 & 0 & 0 & \alpha_1 & \alpha_2 & 0 & \cdots & 0 \\
  \vdots  & \vdots  & \vdots  & \vdots  &\vdots  & \vdots  &\vdots  & \vdots  &\ddots & \vdots  \\
  0 & 0 & 0 & 0 & 0 & 0 & 0 & 0 &  \cdots & \alpha_4 \\
 \end{pmatrix} $

Performing those operations, we can convert any convolution layer into a fully connected layer and then apply the different results found on FC layer.

After briefly introducing the LRP method, we have seen how batch normalization can be theoretically supported by the LRP method. In the next section, several experiments are conducted to analyze the impact of taking the batch normalization into account on both fully connected and convolutional layer. The previous section has also shown how other normalizations can be used when working with LRP. This last part is not studied experimentally.

\section{Experiments}
In this section, we show the results obtained by applying our method to handle batch normalization while using LRP.

Two data-sets are tested with our method. MNIST is an interesting data set for this work, because it is simple enough to achieve good results with fully connected layer but as soon as entrance data are two dimensional images, it can also be treated with a convolutional network.
However MNIST data set is particularly simple, the method is also tested on another data set to consolidate our results. We choose to work with CIFAR-10. Because CIFAR-10 are more complex data, a network composed with fully connected layers will not give usable results. Consequently, only convolutional layers are studied.

Images from both data-sets are normalized such as all pixel values are contained in [-1,1] (see equation \ref{eq:10}).
\begin{equation}\label{eq:10}
    \forall i, \forall j , x_{ij} \in [0, 255], {x_{ij}}_{norm} = \frac{\frac{x_{ij}}{255}-0.5}{0.5}
\end{equation}

% The softmax activation function is used for the last layer. 

% The results of LRP are heat-maps, highlighting pixels with high importance in the classification. One those maps, the more a pixel is red, the more it has been important for the network to take its classification decision.

With toy data sets, the heat-maps are expected to display the same pixels as a human eye would do. A satisfying explainable method creates a heat-map in which the contours and important shapes of objects are intensely red while background elements and insignificant detailed are white.

\citeauthor{Montavon2017b}(\citeyear{Montavon2017b}) bypass normalization layers, it is the baseline we choose to measure our contribution.

% In order to compare the effect of the addition of batch normalization in LRP, to different relevance calculation are performed for networks containing batch normalization layers :
%     \item First, the batch normalization are bypassed and are not taken into account. In the training phase, there is on change but for the relevance propagation, we apply the method just as the batch normalization do not exist.

\subsection{MNIST}
\subsubsection{Fully connected neural network}
\begin{figure}
    \centering
    \includegraphics[width=8cm]{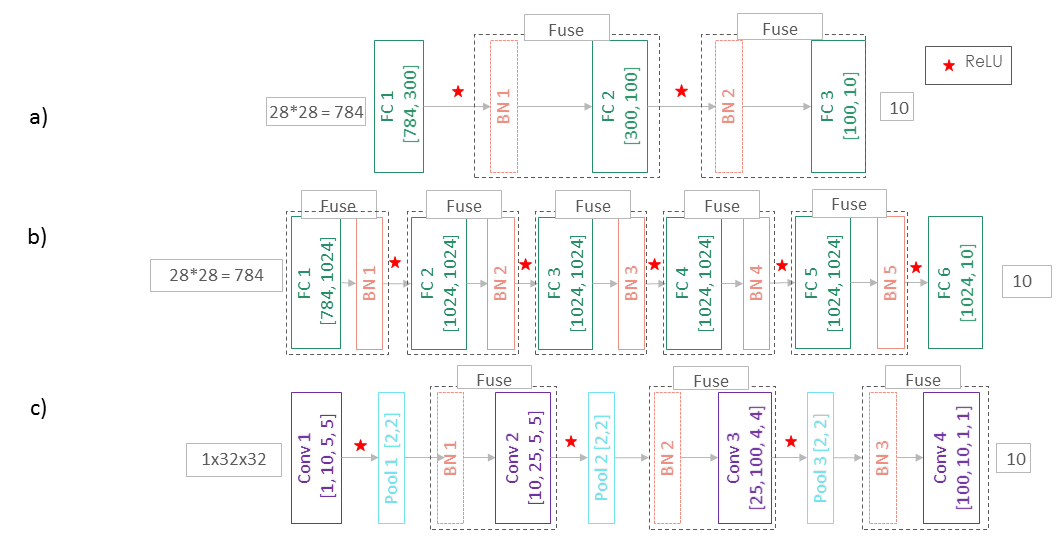}
    \caption{Different networks configurations tested on MNIST data set. The \textbf{a)} figure is the network Fc2 with BN layers before FC layers, Fc1 has the same architecture as Fc2 without BN layers \textbf{b)} is the fully connected layer Fc3 with FC before BN layers. Finally network \textbf{c)} is the architecture of the convolutional network Conv2, Conv1 corresponds to the architecture of conv2 without the BN layers}
    \label{fig:3}
\end{figure}

Three fully connected neural networks only composed with BN and FC layers are studied. Fc1 is only composed with FC layers, Fc2's architecture is similar to Fc1 adding BN layers after activation function. Fc3 presents a little different architecture and uses a BN layer after FC operation and before non linearity. Figure \ref{fig:3} a) and b) detail those networks. 
% \begin{itemize}
%     \item First network Fc1, has three fully connected layers 
%     \item Second network Fc2, has three fully connected layers and two BN layers. It considers batch normalization always before FC layers
%     \item The third network Fc3, is composed of six fully connected layers and five BN layers and FC layers are put before BN ones.
% \end{itemize} 

The networks' performances are measured with the accuracy criterion. All networks give good results, with an accuracy between 97 and 99.24\% (see Table \ref{tab:1}).

\begin{table}
    \centering
    \begin{tabular}{l|l}
        \textbf{Network architecture} & \textbf{Accuracy}\\
       Fc1 - Fully connected without BN  & 0.9781 \\
       Fc2 - Fully connected with BN before FC layers &  0.9742 \\
       Fc3 - Fully connected with BN after FC layers & 0.9831\\
       Conv1 - Convolutional Layer without BN & 0.9903\\
       Conv2 - Convolution Layer with BN & 0.9924\\
    \end{tabular}
    \caption{Accuracy computed on MNIST data set}
    \label{tab:1}
\end{table}

\begin{figure*}
    \includegraphics[width = 18 cm]{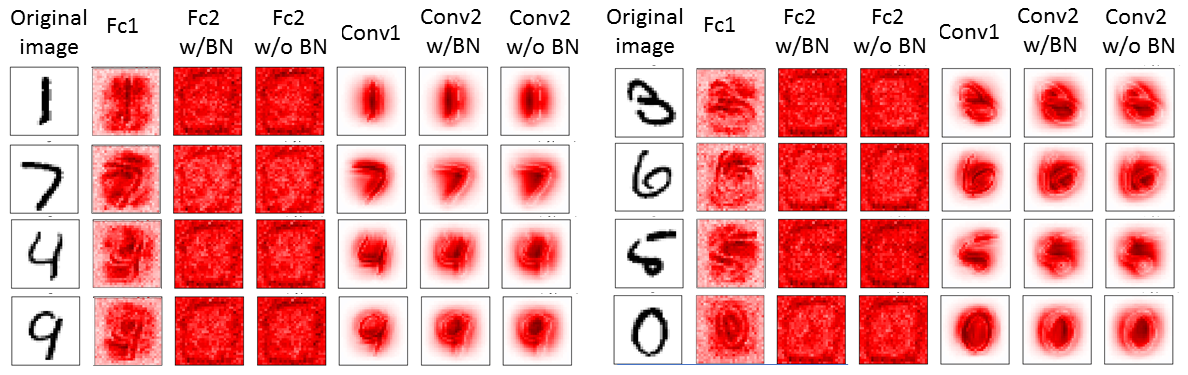}
    \caption{Heat-maps obtained by applying LRP to MNIST data set through different networks. The first column \textbf{Original image} is the raw image, \textbf{Fc1} is the result obtained with the first FC network Fc1,   \textbf{Fc2 w/BN}, is the heat-map obtained with the Fc2 network architecture taking BN layers into account during the LRP phase, on the contrary \textbf{Fc2 w/o BN} column is the heat-map obtained with Fc2 bypassing normalization layers while relevance backpropagation. Similarly, the heatmap resulting from the Conv1 architecture (without BN layers) is given in column \textbf{Conv1}, and columns \textbf{Conv2 w/o BN}  and \textbf{Conv2 w/o BN} show respectively the Conv2 network heat maps result with BN and bypassing BN layers for LRP.}
    \label{fig:4}
\end{figure*}

\subsection{Convolutional neural network}
The procedure applied for the fully connected network is repeated with a convolutional network architecture. Two convolutional networks are built the Conv1 with four convolutional layers, and the Conv2 adding a batch normalization layer before every convolutional layer. Those networks architecture are detailed on figure \ref{fig:3} c).

Some of the heat-maps obtained with LRP method are shown in \ref{fig:4}. Fc2 and Conv2 give two different heat-maps, the first (Figure \ref{fig:4} column 'Fc2 w/ BN' or 'Conv2 w/ BN') using the method developed in this paper, the other (Figure \ref{fig:4} column 'Fc2 w/o BN' or 'Conv2 w/o BN') bypassing the normalization layers i.e. the baseline.

% The application of LRP to Conv1 results as heat-maps displayed on figure \ref{fig:4}, on the 5th column. The Conv2 gives two different heat-maps, the first using the method developed in this paper allowing to include normalization layers in the relevance computation, the other bypassing the normalization layers.

% The only difference in treatment was the inversion between batch normalization layers and convolutional layer which is not study for convolutional layers. 

\subsection{CIFAR-10}
\begin{figure}
    \centering
    \includegraphics[width=8cm]{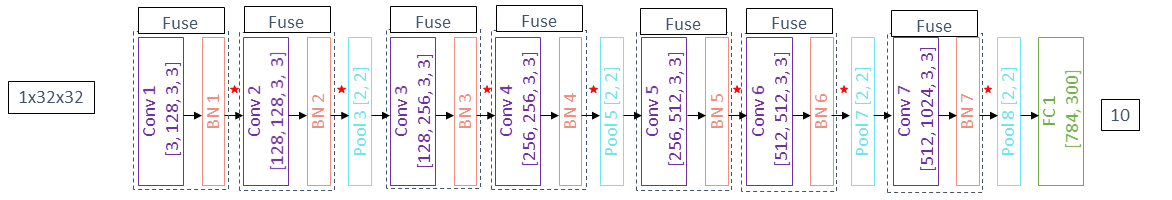}
    \caption{Configuration architecture of the network trained on CIFAR-10 data set}
    \label{fig:5}
\end{figure}

Thechosen  network architecture for the study of CIFAR-10 is composed with seven convolutional layers, each of them is directly followed by a batch normalization layer. It ends with a fully connected layer as it is usually done in classification problems involving convolutional neural networks. Four pooling layers are added to down sampling the intermediate results. More complete information on kernel sizes, and pooling layers location is available in figure \ref{fig:5}.

% Similarly to MNIST data set, results are shown as heat-maps. 

The network reaches an accuracy of 0.9378 on the test set. It leads to two different heat-maps to compare : the first one gives the result with our method meaning considering BN layers while the second one bypasses the BN layers.

\begin{figure}
     \centering
     \includegraphics[width=7cm]{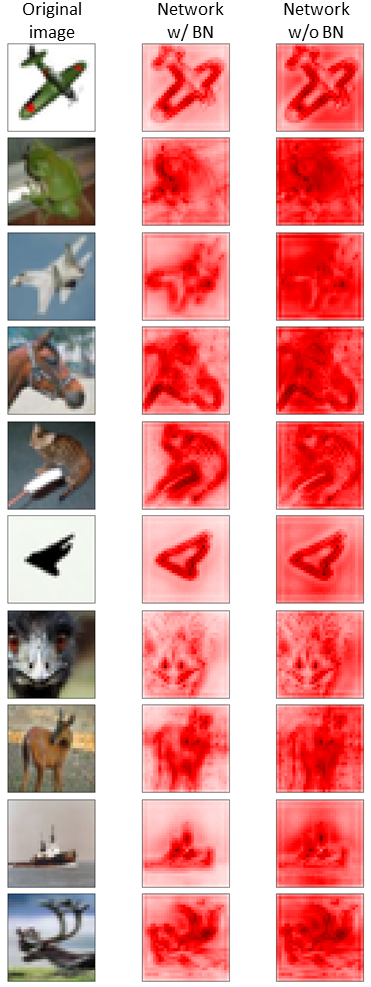}
     \caption{Heat maps showing the entrance pixel relevance in the decision of a convolutional network using the LRP method. The first column \textbf{Original Image} is the image form CIFRA-10 data set studied, the \textbf{Network w/ BN} are results of relevance propagation with the whole model including BN layers, in comparison column \textbf{Network w/o BN} shows the results for the same image still using LRP but bypassing BN layers.}
     \label{fig:6}
 \end{figure}

 Some result examples are shown in figure \ref{fig:6}.
 
This previous section explains experiments run on MNIST and CIFAR-10 data sets with either fully connected or convolutional layers. In the next section, we discuss the results of these experiments, particularly through the interpretation of figures \ref{fig:4} and \ref{fig:6}

\subsection{Results}
% Finding a criterion to assert the quality of a heat maps is quite difficult. Should we consider that the relevance is correct if all pixels of the object to classyfied are recognized or only the pixels contouring the objetc's shape. There is no simple technique to quantitatively evaluate the relevance information. We can think of a measure for the MNIST datasets quite easily but it will not be reproductible for other data sets and so it is not satisfying. 

The analysis presented here is based on qualitative analysis i.e. we compare visually the red intensity difference between pixels of interest and pixels belonging to the background. 

\subsubsection{Better results obtained with convolutional layers than fully connected layers}
Globally, convolutional layers (see Figure \ref{fig:4}, columns "Conv1", "Covn2 w/BN" and "Conv2 w/o BN") give better results than fully connected (see Figure \ref{fig:4}, columns "Fc1", "Fc2 w/BN" and "Fc2 w/o BN") layers. Relevance computed with convolutional networks marks more the difference between background and figures.

\subsubsection{Using BN with fully connected layer provides poor results for the relevance}
Concerning the fully connected network, the network learned without batch normalization layer captures a good relevance information (see Figure \ref{fig:4}, column Fc1). 

The results on the network learned with batch normalization layers are bad considering batch normalization or not during the relevance phase (see figure \ref{fig:4} columns "Fc2 w/BN" and "Fc2 w/o BN"). In this case heat maps highlight all pixels of the image center. The batch normalization interferes in the relevance computation and takes precedence over the figure relevance signal.

For the fully connected network Fc3 where unlike Fc2, batch normalization is placed after FC layers, the explicit results are not given here but are very similar to Fc2 results. 

With BN, whatever the configuration chosen i.e placed before or after activation in the architecture and bypassed during relevance propagation or using our method, results are unusable and no relevant. This might be explained by MNIST data-set, information is always at the same place in the image. LRP method should be used carefully when dealing with fully connected layers combined with batch normalization layers. 

\subsubsection{Relevance obtained with a convolutional network built without BN highlights all pixel of the object while convolutional network with BN highlights contours}
Heat-maps computed for Fc2 bypassing BN or not during relevance propagation are similar, in this section we will focus only on the columns "Conv1" and "Conv2 w/BN" of figure \ref{fig:4}.

About convolutional networks, results are very satisfying and pixels of interest are well localized. However there are differences between the two heat maps obtained by applying LRP with a model without BN and with a model with BN. Looking with attention at columns Conv1 and Conv2 w/ BN of the figure \ref{fig:4}, it appears that the relevance computed from Conv1 gives importance to the pixels composing the figure while using a model with BN, the edges are spotted and the internal pixels are completely white.

\subsubsection{Relevance heat maps computed with our method gives more accurate results than the baseline}
When BN is employed jointly with convolutional network, relevant pixels found with LRP are the edge of the object's shape.

Concerning MNIST data, there are no big differences between the two last columns of figure \ref{fig:4}. But little nuances are observable specifically for figure comporting a loop like 0 or 9, inside the loop, the red color is eased by taking the BN into account during the LRP computation.

For CIFAR-10 (see Figure \ref{fig:5}), on each example, results obtained when BN is not bypassed are significantly better i.e. the contrast between the background and the object is more apparent when BN is introduced in relevance calculation. When the background is uniform and has a very different color than the object, there is an improvement using BN in relevance but the result bypassing BN is already good, this can be observed on the first (plane) and sixth (bird) images.

When the background is uniform but its color is close to the object's color, LRP using BN gives equivalent results to the ones obtained when the background color was more distant. We can observe this on the third image (plane), the fifth (cat), seventh (bird), and ninth (ship) ones. On the contrary, in this case where the difference between background and object is not that clear, not using BN in the relevance leads to medium result. The shape of the object is distinguishable from the rest of the image, but the contrast between the intensity of pixel belonging or not to the object is not pronounced.

For all other examples, when background is not uniform, LRP without using BN gives poor results as for the frog (second image), the horse (fourth image) or deer (last image). The results for LRP taking BN into account is not as good as the previous ones but are much better than relevance when BN is ignored. 

\section{Conclusion}
In this work we propose a method to properly build heat maps with LRP on network containing normalization. From the combination of a fully connected layer or a convolutional layer and a normalization layer we create a new layer on which we can easily apply LRP. We explicit parameters of this new layer for BN used before or after activation. In practice, the method is tested on two toy data sets : MNIST and CIFAR-10. 
Several conclusions emerge from this studies, mainly we show an improvement using our method compared with baseline i.e. bypassing normalization layers. Our study seems to show that the more inputs will be complex, the more benefits achieved with our method will be important. Furthermore, we have noticed that using LRP with a fully connected layer containing BN leads to irrelevant heat-maps in the case of MNIST data-set. There is no proof that this observation is true for all data sets, care must be taken with this configuration.
For future work, other normalization can be tested to evaluate the impact of our method and the improvement provided. The case of fully connected layer should be examined in detail to ensure that this very particular data set is not involved in the bad results. Finally, in this work, we chose to evaluate the heat-maps qualitatively, the development of a method to measure the accuracy of a heat map would give a more reliable comparison between all our results, and might also be a research track.

\bibliography{AAAI20}
\bibliographystyle{aaai}
\end{document}